\documentclass[journal]{IEEEtran}

\usepackage{cs}
\usepackage{mathsymb}
\usepackage{algorithm}
\usepackage{algorithmic}
\usepackage{graphicx}
\usepackage{url}
\usepackage{cases}
\usepackage{subfigure}
\usepackage{diagbox}
\usepackage{times}
\usepackage{epsfig}
\usepackage{amsmath}
\usepackage{amssymb}
\usepackage{multirow}
\usepackage{color}

\begin{document}
\title{Towards End-to-End Car License Plates Detection and Recognition with Deep Neural Networks}

\author{Hui~Li,
        Peng~Wang$^{\dagger}$,
        and~Chunhua~Shen%

\thanks{H. Li, and C. Shen are with the School of Computer Science, The University of Adelaide, SA 5005, Australia. P. Wang is with the School of Computer Science, Northwestern Polytechnical University, China.  $^{\dagger}$P. Wang is the corresponding author (E-mail: peng.wang@nwpu.edu.cn).
This work was done when P. Wang was with The University of Adelaide.
}
}

\markboth{Submitted to IEEE Transactions on Intelligent Transportation Systems 5 April 2017; Revised 26 September 2017 }%
{Li \MakeLowercase{\textit{et al.}}:
	 End-to-End Car License Plates Detection and Recognition}

\maketitle

\begin{abstract}
	In this work, we tackle the problem of car license plate detection and recognition in natural scene images. We propose a unified deep neural network which can localize license plates and recognize the letters simultaneously in a single forward pass. The whole network can be trained end-to-end. In contrast to existing approaches which take license plate detection and recognition as two separate tasks and settle them step by step, our method jointly solves these two tasks by a single network. It not only avoids intermediate error accumulation, but also accelerates the processing speed. For performance evaluation, three datasets including images captured from various scenes under different conditions are tested. Extensive experiments show the effectiveness and efficiency of our proposed approach.

\end{abstract}


\IEEEpeerreviewmaketitle

\section{Introduction}

\IEEEPARstart{A}{utomatic} car license plate detection and recognition plays an important role in intelligent transportation systems. It has a variety of potential applications ranging from security to traffic control, and attracts considerable research attentions during recent years.

However, most of the existing algorithms only work well either under controlled conditions or  with sophisticated image capture systems.
It is still a challenging task to read license plates accurately in an uncontrolled environment.
The difficulty lies in the highly complicated backgrounds, like the general text in shop boards, windows, guardrail or bricks, and random photographing conditions, such as illumination, distortion, occlusion or blurring.

Previous works on license plate detection and recognition usually consider plate detection and recognition as two separate tasks, and solve them respectively by different methods.
However, the tasks of plate detection and recognition are highly correlated. Accurate bounding boxes obtained via detection method can improve recognition accuracy, while the recognition result can be used to eliminate false positives vice versa. Thus in this paper, we propose a unified framework to jointly tackle these two tasks at the same level. A deep neural network is designed, which takes an image as input and outputs the locations of license plates as well as plate labels simultaneously, with both high efficiency and accuracy. We prove that the low level features can be used for both detection and recognition. The whole network can be trained end-to-end, without using any heuristic rule. An overview of the network architecture is shown in Figure~\ref{fig:overview}.  To our knowledge, this is the first work that integrates both license plate detection and recognition into a single network and solves them at the same time. The main contributions of this work are as follows:

\begin{figure*}
	\begin{center}
		\includegraphics[width=0.85\textwidth]{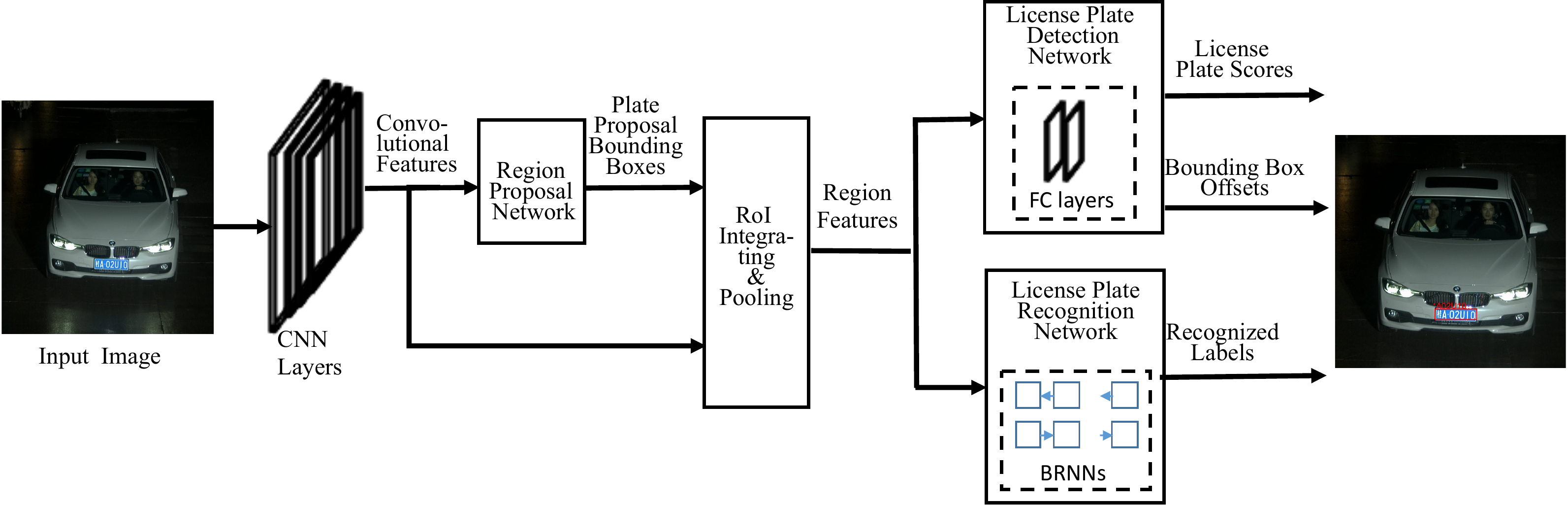}
	\end{center}
	\caption{The overall structure of our model. It consists of several convolutional layers, a region proposal network for license plate proposals generation, proposal integrating and pooling layer, multi-layer perceptrons for plate detection and bounding box regression, and RNNs for plate recognition. Given an input RGB image,  with a single forward evaluation, the network outputs scores of predicted bounding boxes being license plates, bounding box offsets with a scale-invariant translation and log-space height/width shift relative to a proposal, as well as the recognized license plate labels at the same time. The extracted region features are used by both detection and recognition, which not only shares computation, but also reduces model size.}
	\label{fig:overview}
\end{figure*}

\begin{itemize}
	\item
	A single unified deep neural network which can detect license plates from an image and recognize the labels all at once. The whole framework involves no heuristic processes, such as the use of plate colors or character space, and avoids intermediate procedures like character grouping or separation. It can be trained end-to-end, with only the image, plate positions and labels needed for training. The resulting system achieves high accuracy on both plate detection and letter recognition.

	\item
	Secondly, the convolutional features are shared by both detection and recognition, which leads to fewer parameters compared to using separated models. Moreover, with the joint optimization of both detection and recognition losses, the extracted features would have richer information. Experiments show that both detection and recognition performance can be boosted via using the jointly trained model.

	\item
	By integrating plate recognition directly into the detection pipeline, instead of addressing them by separate models, the resulting system is more efficient. With our framework, we do not need to crop the detected license plates from the input image and then recognize them by a separate network. The whole framework takes $0.3 - 0.4$ second per image on a Titan X GPU.

\end{itemize}

The rest of the paper is organized as follows. Section $2$ gives a brief discussion on related work. Section $3$ presents the integrated method, and introduces each part in detail. Experimental verifications are followed in Section $4$, and conclusions are drawn in Section $5$.

\section{Related work}
\label{sec:ReWork}

As license plate detection and recognition are generally addressed separately, we give a brief introduction to previous work on each aspect respectively.

\subsubsection{License Plate Detection} License plate detection aims to localize the license plates in the image in the form of bounding boxes. Existing methods can be roughly classified into four categories~\cite{Du2013Automatic,Zhou2012Principal,Anagnostopoulos}: edge-based,  color-based, texture-based, and character-based.

Since license plates are normally in a rectangular shape with a specific aspect ratio, and they present higher edge density than elsewhere in the image, edge information is used widely to detect license plates. In~\cite{Hsu2013} an edge-based method was developed for plate detection. Expectation Maximization (EM) was applied for edge clustering which extracts the regions with dense sets of edges and with shapes similar to plates as the candidate license plates.
In~\cite{Yuan2017}, a novel line density filter approach was proposed to connect regions with high edge density and remove sparse regions in each row and column from a binary edge image.
Edge-based methods are fast in computation, but they cannot be applied to complex images as they are too sensitive to unwanted edges.

Color-based approaches are based on the observation that color of the license plate is usually different from that of the car body. In~\cite{Ashtari2014}, a plate detection method was developed by analyzing the target color pixels. A color-geometric template was utilized to localize Iranian license plates via strip search. Chang~\etal~\cite{Chang2004} proposed a method to detect Taiwan license plates in RGB images based on the different foreground and background colors. They developed a color edge detector which is sensitive to black-white, red-white and green-white edges. %
Color-based methods can be used to detect inclined or deformed license plates. However, they are very sensitive to various illumination conditions in natural scene images, and they cannot distinguish other objects in the image with similar color and size as the license plates.

Texture-based approaches attempted to detect license plates according to the unconventional pixel intensity distribution in plate regions. Yu~\etal~\cite{Yu2015} used a wavelet transform at first to get the horizontal and vertical details of an image. Empirical Mode Decomposition (EMD) analysis was then employed to deal with the projection data and locate the desired wave crest which indicates the position of a license plate. %
Giannoukos~\etal~\cite{Giannoukos2010} developed a Sliding Concentric Window (SCW) algorithm to identify license plates based on the local irregularity property of plate texture in an image. Operator Context Scanning (OCS) was proposed to accelerate detection speed. %
Texture-based methods use more discriminative characteristics than edge or color, but result in a higher computational complexity.

Considering the fact that license plates consist of a string of characters, much work appeared recently based on the character-based feature as it includes more specific information. Zhou~\etal~\cite{Zhou2012Principal} formulated license plate detection as a visual matching problem. Principal Visual Word (PVW) was generated for each character which contains geometric clues such as orientation, characteristic scale and relative position, and used for plate extraction.
Li~\etal~\cite{Li2013} applied Maximally Stable Extremal Region (MSER) at the first stage to extract candidate characters in images. Conditional Random Field (CRF) was then constructed to represent the relationship among license plate characters. License plates were finally localized through the belief propagation inference on CRF. Character-based methods are more reliable and can lead to a high recall. However, the performance is affected largely by other text in the image background.

\subsubsection{License plate recognition}
Previous work on license plate recognition typically segments characters in the license plate firstly, and then recognizes each segmented character using Optical Character Recognition (OCR) techniques. For example, In ~\cite{Gou2016}, Extremal Regions (ER) were employed to segment characters from coarsely detected license plates and to refine plate location. Restricted Boltzmann machines were applied to recognize the characters. In~\cite{Hsu2013}, MSER was adopted for character segmentation. Local Binary Pattern (LBP) features were extracted and classified using a Linear Discriminant Analysis (LDA) classifier for character recognition.

However, character segmentation by itself is a really challenging task that is prone to be influenced by uneven lighting, shadow and noise in the image. It has an immediate impact on plate recognition. The plate cannot be recognized correctly if the segmentation is improper, even if we have a strong recognizer.
With the development of deep neural networks, approaches were proposed to recognize the whole license plate directly with segmentation free. In~\cite{Bulan2016}, segmentation and optical character recognition were jointly performed using Hidden Markov Models (HMMs) where the most likely label sequence was determined by Viterbi algorithm. In~\cite{Li2016}, plate recognition was regarded as a sequence labeling problem. Convolutional Neural Networks (CNNs) was employed in a sliding window manner to extract a sequence of feature vectors from license plate bounding box. Recurrent Neural Networks (RNNs) with Connectionist Temporal Classification (CTC)~\cite{Graves2009Pami} were adopted to label the sequential data without character separation.

\section{Model}
\label{sec:Detection}

Different from the above-mentioned methods, our approach addresses both detection and recognition using a single deep network. As illustrated in Figure~\ref{fig:overview}, our model consists of a number of convolutional layers to extract discriminate features for license plates, a region proposal network tailored specifically for car license plates, a Region of Interest (RoI) pooling layer, multi-layer perceptrons for plate detection and bounding box regression, and RNNs with CTC for plate recognition. With this architecture, the plate detection and recognition can be achieved simultaneously, with  one network and a single forward evaluation of the input image. Moreover, the whole network is trained end-to-end, with both localization loss and recognition loss being jointly optimized, and shows improved performance.
In the following subsections, we give a detailed description about each component.

\subsection{Model Architecture}

\subsubsection{Low-level Feature Extraction}

The VGG-$16$ network~\cite{Simonyan14c} is adopted here to extract low level CNN features. VGG-$16$ consists of $13$ layers of $3 \times 3$ convolutions followed by Rectified Linear Unit (ReLU) non-linearity, $5$ layers of $2 \times 2 $ max-pooling, and fully connected layers. Here we keep all the convolutional layers and abandon the fully connected layers as we require local features at each position for plate detection. Given that the license plates are small compared with the whole image size, we use $2$ pooling layers instead of $5$, in case the feature information of license plates is vanished after pooling. So the resulting feature maps are one fourth size of the original input image. The higher-resolution feature maps will benefit the detection of small objects~\cite{yolo2017}. They are used as a base for both detection and recognition.

\subsubsection{Plate Proposal Generation}

Ren~\etal~\cite{renNIPS15fasterrcnn} designed a Region Proposal Network (RPN) for object detection, which can generate candidate objects in images.  RPN is a fully convolutional network which takes the low-level convolutional features as input, and outputs a set of potential bounding boxes. It can be trained end-to-end so that high quality proposals can be generated. In this work, we modify RPN slightly to make it suitable for car license plate proposal.

According to the scales and aspect ratios of license plates in our datasets, we designed $6$ scales (the heights are respectively $5$, $8$, $11$, $14$, $17$, $20$) with an aspect ratio ($\text{width}/\text{height}=5$), which results in $k=6$ anchors at each position of the input feature maps. %
In addition, inspired by inception-RPN~\cite{Zhong2016}, we use two $256$-d rectangle convolutional filters ($W_1=5,H_1=3$ and $W_2=3,H_2=1$) instead of the regularly used one filter size $3 \times 3$. The two convolutional filters are applied simultaneously across each sliding position. The extracted local features are concatenated along the channel axis and form a $512$-d feature vector, which is then fed into two separate fully convolutional layers for plate/non-plate classification and box regression. On one hand, these rectangle filters are more suitable for objects with larger aspect ratios (license plates). On the  other hand, the concatenated features keep both local and contextual information, which will benefit the plate classification.

For $k$ anchors at each sliding position on the feature map, the plate classification layer outputs $2k$ scores which indicate the probabilities of the anchors as license plates or not. The bounding box regression layer outputs $4k$ values which are the offsets of anchor boxes to a nearby ground-truth. Given an anchor with the center at $({x_a}, {y_a})$, width $w_a$ and height $h_a$, the regression layer outputs $4$ scalars $(t_x, t_y, t_w, t_h)$ which are the scale-invariant translation and log-space height/width shift. The bounding box after regression is given by
\begin{align*}
x=x_a + t_x w_a, y=y_a + t_y h_a,
\\ \
w=w_a \exp(t_w), h=h_a \exp(t_h),
\end{align*}
where $x,y$ are the center coordinates of the bounding box after regression, and $w,h$ are its width and height.

For a convolutional feature map with size $M \times N$, there will be $M \times N \times k$ anchors in total. Those anchors are redundant and highly overlapped with each other. Moreover, there are much more negative anchors than positive ones, which will lead to bias during training if we use all those anchors. We randomly sample 256 anchors from one image as a mini-batch, where the ratio between positive and negative anchors is up to 1:1. The anchors that have Intersection over Union (IoU) scores larger than $0.7$ with any ground-truth bounding box are selected as positives, while anchors with IoU lower than $0.3$ as negatives. The anchors with the highest IoU scores are also regarded as positives, so as to make sure that every ground-truth box has at least one positive anchor.  If there are not enough positive anchors, we pad with negative ones.

The binary logistic loss is used here for box classification, and smooth $L_1$ loss~\cite{renNIPS15fasterrcnn} is employed for box regression. The multi-task loss function used for training RPN is
\begin{equation}\label{eq1}
L_{{R\!P\!N}}=\frac{1}{N_{cls}} \sum_{i=1}^{N_{cls}} {L}_{{cls}} (p_i,p_i^\star)  + \frac{1}{N_{reg}}
\sum_{i=1}^{N_{reg}} L_{{reg}} (\mathbf{t}_i,\mathbf{t}_i^\star),
\end{equation}
where $N_{cls}$ is the size of a mini-batch and $N_{reg}$ is the number of positive anchors in this batch. Bounding box regression is only for positive anchors, as there is no ground-truth bounding box matched with negative ones.
$p_i$ is the predicted probability of anchor $i$ being a license plate and $p_i^\star$
is the corresponding ground-truth label ($1$ for positive anchor, $0$ for negative anchor).
$\mathbf{t}_i$ is the predicted coordinate offsets $(\mathbf{t}_{i,x}, \mathbf{t}_{i,y}, \mathbf{t}_{i,w}, \mathbf{t}_{i,h})$ for anchor $i$,
and $\mathbf{t}_i^\star$ is the associated offsets for anchor $i$ relative to the ground-truth.  RPN is trained end-to-end with back-propagation and Stochastic Gradient Descent (SGD).

At test time, the forward evaluation of RPN will result in $M \times N \times k$ anchors with objectiveness scores as well as bounding box offsets. We employ Non-Maximum Suppression (NMS) to select $100$  proposals with higher confidences based on the predicted scores for the following processing.

\subsubsection{Proposal Processing and Pooling}

As we state before, $256$ anchors are sampled from the $M \times N \times k$ anchors to train RPN. After bounding box regression, the $256$ samples will later be used for plate detection and recognition.

We denote the bounding box samples as ${p}=({x^{(1)}}, {y^{(1)}}, {x^{(2)}}, {y^{(2)}})$, where $({x^{(1)}},{y^{(1)}})$ is the top-left coordinate of the bounding box, and  $({x^{(2)}},{y^{(2)}})$ is the bottom-right coordinate of the bounding box.
For all the positive proposals ${p_{i,j}}=({x^{(1)}_{i,j}}, {y^{(1)}_{i,j}}, {x^{(2)}_{i,j}}, {y^{(2)}_{i,j}})$, $i=1,\dots, n$ that are associated with the same ground truth plate ${g_j}$, a bigger bounding box ${b_j}=({x^{(1)}_{j}}, {y^{(1)}_{j}}, {x^{(2)}_{j}}, {y^{(2)}_{j}})$ is constructed that encompasses all proposals ${p_{i,j}}$, \ie,
\begin{align*}
x^{(1)}_{j}=\min_{i=1,\dots,n}(x^{(1)}_{i,j}), && y^{(1)}_{j}=\min_{i=1,\dots,n}(y^{(1)}_{i,j}),
\\ \
x^{(2)}_{j}=\max_{i=1,\dots,n}(x^{(2)}_{i,j}), &&   y^{(2)}_{j}=\max_{i=1,\dots,n}(y^{(2)}_{i,j}).
\end{align*}

The constructed bounding boxes ${b_j}$, $j=1, \dots, m$ will then be used as positive samples for later plate detection and recognition.  To avoid the bias caused by the unbalanced distribution between positive and negative samples, we randomly choose $3m$ negative ones from the $256$ samples and form a mini-batch with $4m$ samples.

Considering that the sizes of the samples are different from each other, in order to interface with the plate detection network as well as the recognition network, RoI pooling~\cite{RGB2015ICCV} is adopted here to extract fixed-size feature representation.  Each  RoI is projected into the image convolutional feature maps, and results in feature maps of size $H' \times W'$. The varying sized feature maps $H' \times W'$ are then divided into $X \times Y$ grids, where boundary pixels are aligned by rounding. Features are max-pooled within each grid. Here we choose $X=4$ and $Y=20$ instead of $7 \times 7$ that is used in~\cite{RGB2015ICCV}, because of the subsequent plate recognition task. To be specific, since we need to recognize each character in the license plate, it would be better if we keep more feature horizontally. However, the model size $\emph{p}$ from this layer to the next fully connected layer is closely related to $X$ and $Y$, \ie, $\emph{p} \propto {XY}$. A larger feature map size will result in more parameters and increase the computation burden. Considering the aspect ratio of license plates, we use a longer width $Y=20$ and a shorter height $X=4$. Experimental results demonstrate that these features are sufficient for classification and recognition.

\subsubsection{Plate Detection Network}

Plate detection network aims to judge whether the proposed RoIs are car license plate or not, and refine the coordinates of plate bounding boxes.

Two fully connected layers with $2048$ neurons and a dropout rate of $0.5$ are employed here to extract discriminate features for license plate detection. The features from each RoI are flattened into a vector and passed through the two fully connected layers.
The encoded features are then fed concurrently into two separate linear transformation layers respectively for plate classification and bounding box regression. The plate classification layer has $2$ outputs, which indicate the softmax probability of each RoI as plate/non-plate. The plate regression layer produces the bounding box coordinate offsets for each proposal, as in region proposal network.

\subsubsection{Plate Recognition Network}

\begin{figure}
	\begin{center}
		\includegraphics[width=0.25\textwidth]{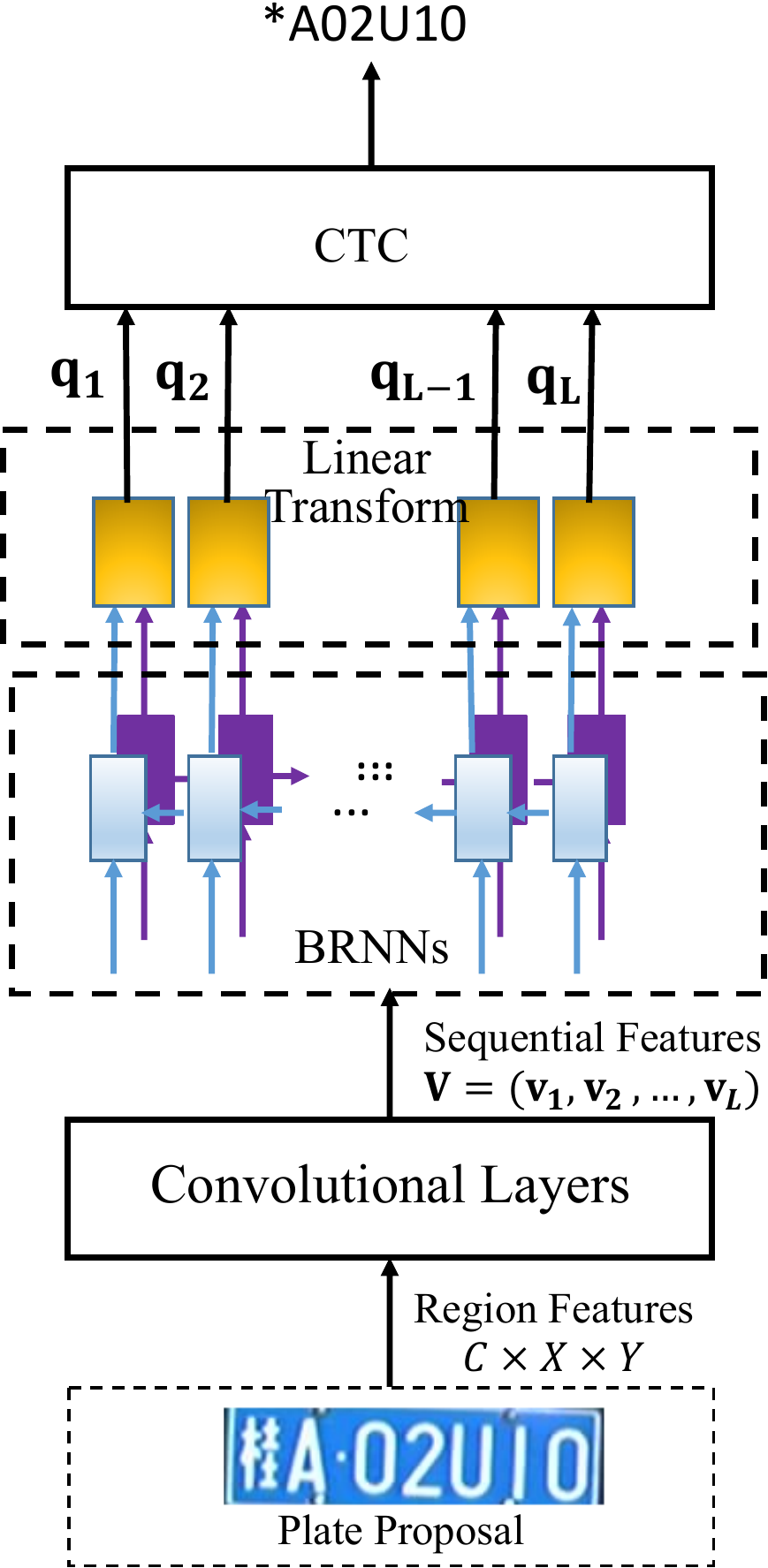}
	\end{center}
	\caption{Plate Recognition Network. The pooled region features are regarded as a feature sequence, and encoded by BRNNs, which capture the context information in both sides. CTC are used for plate decoding without character separation.}
	\label{fig:recognition}
\end{figure}

Plate recognition network aims to recognize each character in RoIs based on the extracted region features. To avoid the challenging task of character segmentation, we regard the plate recognition as a sequence labeling problem. Bidirectional RNNs (BRNNs) with CTC loss~\cite{Graves2009Pami} are employed to label the sequential features, which is illustrated in Figure~\ref{fig:recognition}.

The region features after RoI pooling are denoted as $\mathbf{Q} \in \mathbb{R}^{C \times X \times Y}$, where $C$ is the channel size. First of all, we add two additional convolutional layers with ReLUs. Both of them use $512$ filters. The kernel sizes are $3$ and $2$ respectively, with a padding of $1$ used in the first convolutional layer. %
A rectangular pooling window with $k_W=1$ and $k_H=2$ is adopted between them, which would be beneficial for recognizing characters with narrow shapes, such as ’$1$’ and ’I’, referring to~\cite{ShiBY15}. These operations will reform the region features $\mathbf{Q}$ to a sequence with the size as $D \times L$, where  $D=512$ and $L=19$. We denote the resulting features as $\mathbf{V}=(\mathbf{v}_{1}, \mathbf{v}_{2},  \dots, \mathbf{v}_{L})$, where $\mathbf{v}_{i} \in  \mathbb{R}^D$.

Then BRNNs are applied on top of the sequential features. As presented in Figure~\ref{fig:recognition}, Two separated RNN layers with $512$ units are used. One processes the feature sequence forward, with the hidden state updated via $\mathbf{h}^{(f)}_t = \mathrm{g}(\mathbf{v}_t, \mathbf{h}^{(f)}_{t-1})$. The other one processes it backward with the hidden state updated via $\mathbf{h}^{(b)}_t = \mathrm{g}(\mathbf{v}_t, \mathbf{h}^{(b)}_{t+1})$. The two hidden states are concatenated together and fed to a linear transformation with $37$ outputs. Softmax layer is followed to transform the $37$ outputs into probabilities, which correspond to the distributions over $26$ capital letters, $10$ digits, and a special non-character class. %
We record the probabilities at each time step. Hence, after BRNNs encoding, the feature sequence $\mathbf{V}$ is transformed into a sequence of probability estimation $\mathbf{q}=(\mathbf{q}_{1}, \mathbf{q}_{2},  \dots, \mathbf{q}_{L})$ with the same length as $\mathbf{V}$. BRNNs capture abundant contextual information from both directions, which will make the character recognition more accurate. To overcome the shortcoming of gradient vanishing or exploding during traditional RNN training, Long-Short Term Memory (LSTM)~\cite{LSTM} is employed here. It defines a new cell structure called memory cell, and three multiplicative gates (\ie, input gate, forget gate and output gate), which can selectively store information for a long time.

Then CTC layer~\cite{Graves2009Pami} is adopted here for sequence decoding, which is to find an approximately optimal path $\mathbf{\pi}^*$ with maximum probability through the BRNNs' output sequence $\mathbf{q}$, \ie,
\begin{equation}
\mathbf{\pi}^* \thickapprox \mathcal{B} \left(\arg \max_{\mathbf{\pi}} P(\mathbf{\pi} | \mathbf{q})\right).
\end{equation}
Here a path $\mathbf{\pi}$ is a label sequence based on the output activation of BRNNs, and
$P(\mathbf{\pi} | \mathbf{q}) = \prod_{t=1}^{L}P(\mathbf{\pi}_t|\mathbf{q}) $. The operator $\mathcal{B}$ is defined as the operation of removing the repeated labels and the non-character label from the path. For example, $\mathcal{B}(a-a-b-)=\mathcal{B}(-aa--a-bb)=(aab)$.
Details of CTC can refer to~\cite{Graves2009Pami}. The optimal label sequence $\mathbf{\pi}^*$ is exactly the recognized plate label.

\subsection{Loss Functions and Training}

As we demonstrate previously, the whole network takes as inputs an image, the plate bounding boxes and the associated labels during training time. After we obtain the samples as well as the region features, we combine the loss terms for plate detection and recognition, and jointly train the detection and recognition networks. Hence, the multi-task loss function is defined as

\begin{align}
\label{eq2}
L_{D\!R\!N} &= \frac{1}{N} \sum_{i=1}^N L_{cls} (p_i,p_i^\star)
+ \frac{1}{N_{+}} \sum_{i=1}^{N_+} L_{reg} (\mathbf{t}_i,\mathbf{t}_i^\star)  \notag \\
&+ \frac{1}{N_{+}} \sum_{i=1}^{N_+} L_{rec} (\mathbf{q}^{(i)}, \mathbf{s}^{(i)})
\end{align}
where $N$ is the size of a mini-batch used in detection network and $N_+$ is the number of positive samples in this batch. The definitions of $L_{cls}$ and  $L_{reg}$ are the same as that used in RPN. $p_i,p_i^\star,\mathbf{t}_i,\mathbf{t}_i^\star$ also use the same definition as that used in RPN.  $\mathbf{s}^{(i)}$ is the ground truth plate label for sample $i$ and $\mathbf{q}^{(i)}$ is the corresponding output sequence by BRNNs.

It is observed that the length of BRNNs' outputs $\mathbf{q}^{(i)}$ is not consistent with the length of target label $\mathbf{s}^{(i)}$. Following CTC loss in~\cite{Graves2009Pami}, the objective function for plate recognition is defined as the negative log probability of the network outputting correct label, \ie,

\begin{equation}
L_{rec} (\mathbf{q}^{(i)}, \mathbf{s}^{(i)}) = - \log P(\mathbf{s}^{(i)} | \mathbf{q}^{(i)})
\end{equation}

where
\begin{equation}
P(\mathbf{s}^{(i)} | \mathbf{q}^{(i)}) = \sum_{\mathbf{\pi}:  \mathcal{B} (\mathbf{\pi})=\mathbf{s}^{(i)}} P(\mathbf{\pi} | \mathbf{q}^{(i)})
\end{equation}
which is the sum of probabilities of all $\mathbf{\pi}$ that can be mapped to $\mathbf{s}^{(i)}$ by $ \mathcal{B}$.

We use the approximate joint training process~\cite{renNIPS15fasterrcnn} to train the whole network, ignoring the derivatives with respect to the proposed boxes' coordinates. Fortunately, this does not have a great influence on the performance~\cite{renNIPS15fasterrcnn}. We train the whole network using SGD. CNNs for extracting low-level features are initialized from the pre-trained VGG-$16$ model. We do not fine-tune the first four convolutional layers for efficiency. The rest of CNN layers are fine-tuned only in the first $50K$ iterations. The other weights are initialized according to Gaussian distribution. For optimization, we use ADAM~\cite{adam14}, with an initial learning rate of $10^{-5}$ for parameters in the pre-trained VGG-$16$ model, and $10^{-4}$ for other parameters. The latter learning rate is halved every $10K$ iterations until $10^{-5}$. The network is trained for $200K$ iterations. Each iteration uses a single image sampled randomly from training dataset. For each training image, we resize it to the shorter side of $700$ pixels, while the longer side no more than $1500$ pixels.

\section{Experiments}
\label{SEC:Exp}
In this section, we conduct experiments to verify the effectiveness of the proposed methods.  Our network is implemented using Torch 7. The experiments are performed on NVIDIA Titan X GPU with $12$GB memory.

\subsection{Datasets}
Three datasets are used here to evaluate the effectiveness of our proposed method.

The first dataset is composed of car license plates from China, denoted as "CarFlag-Large". We collected $450K$ images for training, and $7378$ images for test. The images are captured from frontal viewpoint by fixed surveillance cameras under different weather and illumination conditions, \eg, in sunny days, in rainy days, or at night time, with a resolution of $1600 \times 2048$. The plates are nearly horizontal. Only the nearest license plate in the image is labeled in the ground truth file.

The second dataset is the Application-Oriented License Plate (AOLP) database~\cite{Hsu2013}. It has $2049$ images in total with Taiwan license plates. This database is categorized into three subsets with different level of difficulty and photographing condition, as refer to ~\cite{Hsu2013}: Access Control (AC), Traffic Law Enforcement (LE), and Road Patrol (RP). Since we do not have any other images with Taiwan license plates, to train the network, we use images from different sub-datasets for training and test separately. For example, we use images from LE and RP subsets to train the network, and evaluate the performance on AC subset. Considering the small number of training images, data augmentation is implemented by rotation and affine transformation.

The third dataset is issued by Yuan~\etal~\cite{Yuan2017}, and denoted as "PKUData". It has $3977$ images with Chinese license plates captured from various scenes. It is categorized into $5$ groups (\ie, G$1$-G$5$) corresponding to different configurations, as introduced in~\cite{Yuan2017}. However, there are only the plate bounding boxes given in the ground truth file. Hence, we merely evaluate the detection performance on this dataset.

\begin{figure*}[h]
	\begin{center}
		\includegraphics[width=0.95\textwidth]{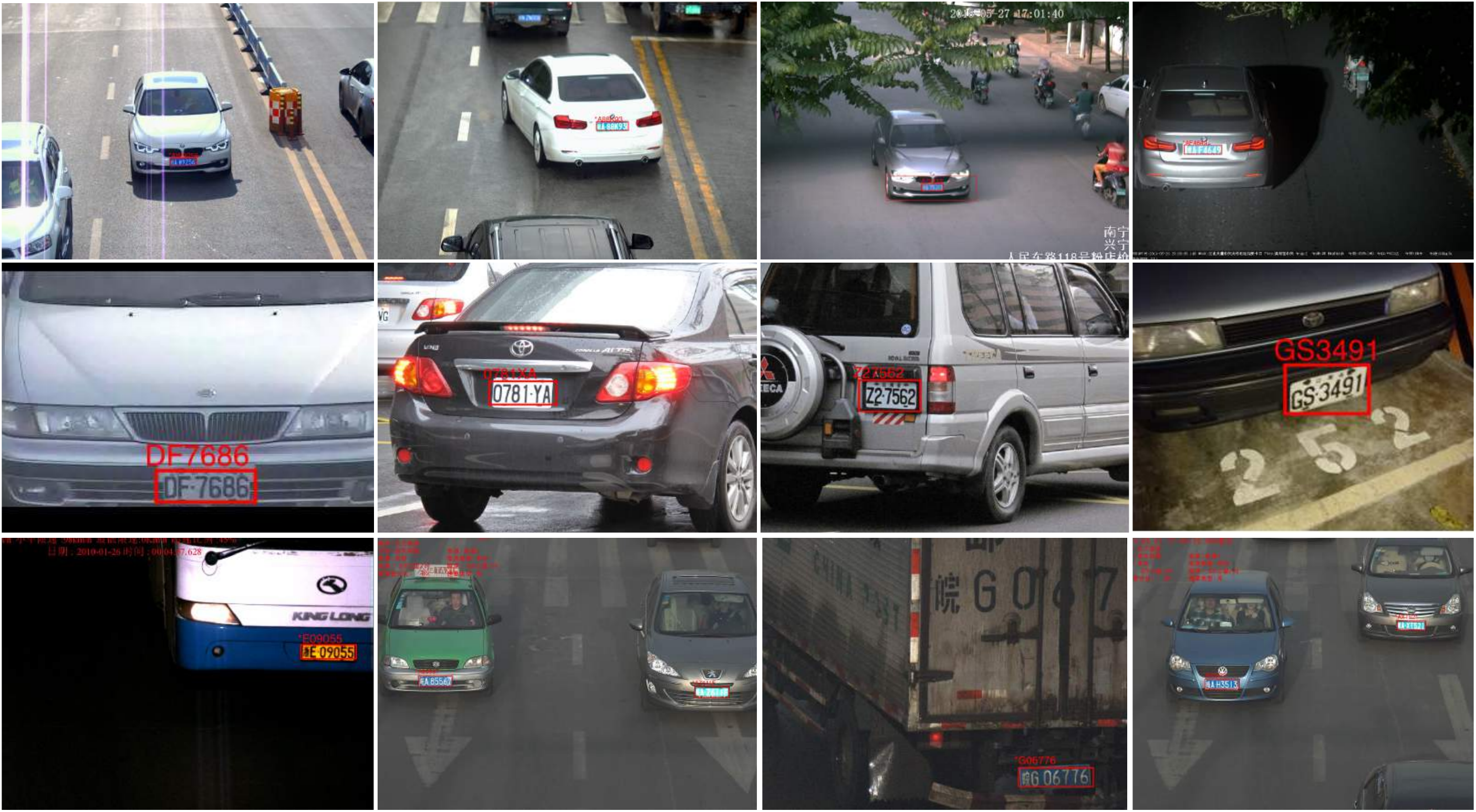}
	\end{center}
	\caption{Example results for open wide car license plate detection and recognition by our jointly trained model. Images in the first line are from CarFlag-Large, the second line are from AOLP and the third line are from PKUData. The results demonstrate that our model can detect and recognize car license plates under various photographing conditions, such as day and night, sunny and rainy days, \etc. }
	\label{fig:result}
	\vspace{-0.3cm}
\end{figure*}

\subsection{Evaluation Criterion}
To evaluate the ``End-to-end'' performance with both detection and recognition results considered, we follow the "End-to-end" evaluation protocol for general text spotting in natural scene~\cite{icdar2015} as they have similar application scenario. Define IoU as
\begin{eqnarray}
\text{IoU} = \frac{\mbox{area}(R_{\mbox{det}} \cap R_{\mbox{gt}}) }{\mbox{area}(R_{\mbox{det}} \cup R_{\mbox{gt}})}
\end{eqnarray}
where $R_{\mbox{det}}$ and $R_{\mbox{gt}}$ are regions of the detected bounding box and ground-truth respectively.

The bounding box is considered to be correct if its IoU with a ground truth bounding box is more than $50\%$ ($\text{IoU} > 0.5$), and the plate labels match. It should be note that we denote all Chinese character in license plates as `*', since the training images in CarFlag-Large are all from one province and use the same Chinese character. The trained network can not be used to distinguish other Chinese characters.

As to the detection-only performance, we follow the criterion used in~\cite{Yuan2017} for fair competition, \ie, a detection is considered to be correct if the license plate is totally encompassed by the bounding box, and $\text{IoU} > 0.5$. %

\subsection{Performance Evaluation on CarFlag-Large}
\begin{figure}
	\begin{center}
		\includegraphics[width=0.5\textwidth]{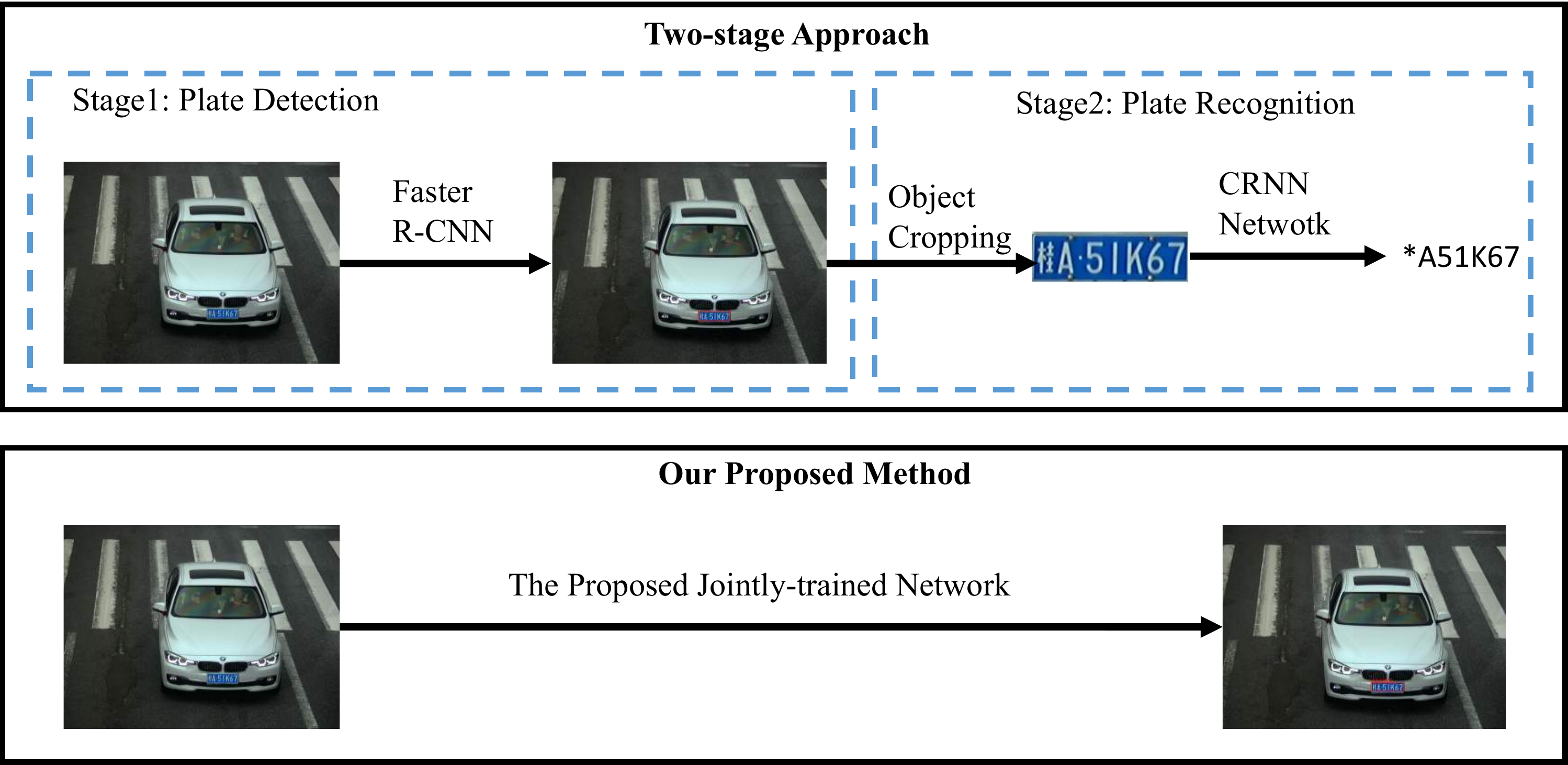}
	\end{center}
	\caption{Two-stage approach VS. our proposed method. In the two-stage approach, after license plate detection by Faster R-CNN, we crop the detected license plates from the image, and then recognize them by another separate model (CRNN in this paper). The features need to be re-computed during recognition phase. In contrast our proposed network takes an image as input, and produces license plate bounding boxes and plate labels in one-shot. It avoids some intermediate processes like image cropping, and share computation for convolutional feature extraction.}
	\label{fig:comp}
	\vspace{-0.3cm}
\end{figure}

In this section, we would like to demonstrate the superiority of our end-to-end jointly trained framework compared with commonly used two-stage approaches. As illustrated in Figure~\ref{fig:comp}, a commonly used two-stage approach implements plate detection and recognition by two separated models. Plate detection is carried out firstly. The detected objects are cropped out and then recognized by another different model. In contrast, our proposed network outputs both detection and recognition results at the same time, with a single forward pass and requiring no image cropping. The convolutional features are shared by both detection and recognition, which omits feature re-computation. For simplicity, we denote our jointly trained network as ``Ours (Jointly-trained)'', and the two stage approach as ``Ours (Two-stage)''. The model used only for plate detection is denoted as ``Ours (Detection-only)''.

For fair competition, we train a Faster R-CNN~\cite{renNIPS15fasterrcnn} model using the $450K$ training images for plate detection only. We modify the scales and shapes of anchors as the ones we used in this paper so that they fit the license plates. The network is also trained with $200K$ iterations, using the same initial parameters and learning rate.  As to the plate recognition, we employ CRNN framework~\cite{ShiBY15}, which produces the state-of-the-art performance on general text recognition. It is an end-to-end framework for cropped word recognition, including CNN layers, RNN layers and CTC for transcription, from bottom to top. We crop the groud-truth license plates from the $450000$ training images, and resize them to $160 \times 32$ pixels. Then we fine-tune the CRNN model with these training data.

In order to boost the performance, we rescale the input image into multiple sizes during test phase for both our proposed network and the detection-only Faster R-CNN network. The input images are resized to the shorter side of $600$, $1200$ pixels respectively, while the longer side less than $1500$ pixels. With our framework, both detection and recognition results come out  together, while with the two-stage approach, we crop the detected bounding boxes from input images, resize them to $160 \times 32$ pixels, and then feed into the trained CRNN model for recognition. Only bounding boxes with classification score larger than $0.95$ are kept and merged via NMS. Considering that there is only one plate labeled as ground truth per image, we finally choose the one that has $7$ characters recognized and/or with the highest detection score for evaluation. The test results are presented in Table~\ref{Tab:2}. Our jointly trained network gives the ``End-to-end'' performance with F-measure of $96.13\%$ on $7378$ test images. It is around $2\%$ higher than the results by the two-stage approach, which demonstrates the advantage of end-to-end training for both detection and recognition in an unified network. The learned features are more informative, and the two subtasks can help with each other.

\begin{table}[h]
	\
	\newcommand{\tabincell}[2]{\begin{tabular}{@{}#1@{}}#2\end{tabular}}
	\begin{center}
		\caption{Experimental results on CarFlag-Large dataset. We compare both performance and running speed of our jointly trained network with a two-stage baseline method. The jointly trained network achieves not only higher accuracies on both detection and ``End-to-end'' performance, but also in a shorter time. }
		\label{Tab:2}
		{
			\begin{tabular}{l|c|c|c}
				\hline
				Method  & \tabincell{c}{End-to-end  \\  Performance \\ (\%)}  & \tabincell{c}{Detection-only  \\  Performance \\ (\%)}   & \tabincell{c}{Speed \\  (per image \\ single scale) \\ (ms)} \\
				\hline
				Ours(Jointly-trained) & $\textbf{96.13}$  & $\textbf{98.15}$ & $\textbf{300}$  \\
				\hline
				Ours(Two-stage)  & $94.09$  & $97.00$ & $450$   \\
				\hline
			\end{tabular}
		}

	\end{center}
\end{table}

In terms of the computational speed, the unified framework takes about $300ms$ per image for a forward evaluation on the single small input scale, while the two-stage approach needs around $450ms$ to get both detection and recognition results, as it needs to implement image cropping and CNN feature re-calculation.

We also compare the detection-only performance. Our jointly trained network produces a detection accuracy of $98.15\%$, which is $1\%$ higher than the result given by detection-only Faster R-CNN network. This result illustrates that car license plate detection can be improved with the multi-task loss used during training time.

Some experimental results using our jointly trained network are presented in the first row of Figure~\ref{fig:result}, which show that our model can deal with images under different illumination conditions.

\subsection{Performance Evaluation on AOLP}
In this section, we compare the ``End-to-end'' performance of our method with other state-of-the-art methods on the  AOLP dataset. Note that the network is only trained with $15K$ iterations because of the small number of training images in this dataset. Moreover, since the sizes of license plates in AOLP are almost the same, and the ratios between license plates and images sizes are also similar. For this dataset, we only use a single image scale with shorter side as $700$ pixels in test phase.

The detection and recognition results are presented on the second row in Figure~\ref{fig:result}. Comparison results with other methods in Table~\ref{Tab:3} show that our approach performs better on AC and LE subsets with ``End-to-end'' evaluation. It also gives the best performance for plate detection on all three subsets, with averagely $2\%$ higher than the sliding window based method used in Li~\etal~\cite{Li2016}, and $4\%$ higher than the edge based method used in Hsu~\etal~\cite{Hsu2013}. As to the computational speed, out network takes about $400ms$ to get both detection and recognition results, while Li~\etal's method ~\cite{Li2016} costs $2-3s$, and Hsu~\etal's approach~\cite{Hsu2013} needs averagely $260ms$.

It should be noted that in Table~\ref{Tab:3}, ``End-to-end'' performance on RP subset is worse than that in~\cite{Li2016}. That may be because the license plates in RP have a large degree of rotation and projective orientation. In~\cite{Li2016}, the detected license plates are cropped out and Hough transform is employed to correct the orientation. In contrast, our method does not explicitly handle the rotated plates. Integrating spatial transform network into our end-to-end framework may be a solution, referring to~\cite{shiCVPR2016}, which is a future work.

\begin{table*}[h]
	\
	\newcommand{\tabincell}[2]{\begin{tabular}{@{}#1@{}}#2\end{tabular}}
	\begin{center}
		\caption{Experimental results on AOLP dataset.  AC (Access Control) is the easiest dataset where images are captured when vehicles pass a fixed passage with a lower speed or full stop. LE (Law Enforcement) dataset consists of images captured by roadside camera when a vehicle violates traffic laws. RP (Road Patrol) refers to the cases that the camera is held on a patrolling vehicle, and the images are taken with arbitrary viewpoints and distances. We compare our proposed method with other state-of-the-art methods on both performance and running speed. Our jointly-trained network shows improved performance for images with license plates in nearly horizontal position. }
		\label{Tab:3}
		{
			\begin{tabular}{l|c|c|c|c|c|c|c}
				\hline
				\multirow{2}{*}{Method} & \multicolumn{3}{c}{\tabincell{c}{End-to-end  Performance \\ (\%)} } &  \multicolumn{3}{|c} {\tabincell{c}{Detection-only Performance \\ (\%)} } &  \multicolumn{1}{|c}{ \tabincell{c}{Speed \\  (per image single scale) \\ (ms)}} \\ \cline{2-7}	& \multicolumn{1}{c|}{AC} & \multicolumn{1}{c|}{LE}  & \multicolumn{1}{c|}{RP}   & \multicolumn{1}{c|}{AC} & \multicolumn{1}{c|}{LE} & \multicolumn{1}{c|}{RP} &   \\
				\hline
				Hsu~\etal~\cite{Hsu2013} & $-$  & $-$ & $-$ & $96$ & $95$ & $94$ & $\textbf{260}$ \\
				\hline
				Li~\etal~\cite{Li2016} & $94.85$  & $94.19$ & $\textbf{88.38}$ & $98.38$ & $97.62$ & $95.58$ & $1000-2000$    \\
				\hline
				Ours(Jointly-trained) & $\textbf{95.29}$  & $\textbf{96.57}$ & $83.63$ & $\textbf{99.56}$ & $\textbf{99.34}$ & $\textbf{98.85}$ & $400$    \\
				\hline
			\end{tabular}
		}
		\vspace{-0.6cm}
	\end{center}
\end{table*}

\subsection{Performance Evaluation on PKUData}

\begin{table*} [h]
	\
	\newcommand{\tabincell}[2]{\begin{tabular}{@{}#1@{}}#2\end{tabular}}
	\begin{center}
		\caption{Experimental results on PKUData.  Detection performance and running speed are compared between our proposed method and other state-of-the-art methods. G$1$ - G$5$  correspond to different image capturing conditions. Our jointly trained network achieves a average detection ratio of $99.80\%$, which is $2\%$ higher than the previous best performance method. In addition, the jointly trained network, which integrates both detection and recognition losses, performs better than that trained only with the detection information. }
		\label{Tab:4}
		{
			\begin{tabular}{l|c c c c c|c|c}
				\hline
				\multirow{2}{*}{Method} &  \multicolumn{6}{c} {\tabincell{c}{Detection Performance (\%)} } &  \multicolumn{1}{|c}{ \tabincell{c}{Speed \\  (per image single scale) \\ (ms)}} \\ \cline{2-7}	& \multicolumn{1}{c}{G$1$} & \multicolumn{1}{c}{G$2$}  & \multicolumn{1}{c}{G$3$}   & \multicolumn{1}{c}{G$4$} & \multicolumn{1}{c|}{G$5$} & \multicolumn{1}{c|}{Average} &    \\
				\hline
				Zhou~\etal~\cite{Zhou2012Principal} & $95.43$  & $97.85$ & $94.21$ & $81.23$ & $82.37$ & $90.22$ & $475$ \\
				\hline
				Li~\etal~\cite{Li2013} & $98.89$  & $98.42$ & $95.83$ & $81.17$ & $83.31$ & $91.52$ & $672$    \\
				\hline
				Yuan~\etal~\cite{Yuan2017} & $98.76$  & $98.42$ & $97.72$ & $96.23$ & $97.32$ & $97.69$ & $42$    \\
				\hline
				Ours(Detection-only) & $\textbf{99.88}$  & $99.71$ & $\textbf{99.87}$ & $99.65$ & $98.81$ & $99.58$ & $\textbf{300}$    \\
				\hline
				Ours(Jointly-trained) & $\textbf{99.88}$  & $\textbf{99.86}$ & $\textbf{99.87}$ & $\textbf{100}$ & $\textbf{99.38}$ & $\textbf{99.80}$ & $\textbf{300}$    \\
				\hline
			\end{tabular}
		}

	\end{center}
\end{table*}

Because the ground truth file in PKUData only provides the plate bounding boxes, we simply evaluate the detection performance on this dataset. Both the detection accuracy and computational efficiency are compared with other methods~\cite{Li2013,Zhou2012Principal,Yuan2017}. We use the same model trained by the CarFlag-Large dataset, as they are both datasets with Chinese license plates. %

Images on the third line of Figure~\ref{fig:result} show examples with both detection and recognition results. The detection-only results by our approach and other three methods are presented in Table~\ref{Tab:4}. Our jointly trained model demonstrates absolute advantage on all $5$ subsets, especially on G$4$, where we achieve $100\%$ detection rate. This result proves the robustness of our approach in face of various scenes and diverse conditions. Qualitatively, our jointly trained network achieves a average detection ratio of $99.80\%$, which is $2\%$ higher than the previous best performance method.

In addition, the detection performance by our jointly trained network is slightly better than that by the detection-only network as seen from Table~\ref{Tab:4}. This is consistent with the outcome on CarFlag-Large dataset, and proves again that the detection performance can be boosted when training with the label information.

In terms of computational speed, Yuan~\etal's method~\cite{Yuan2017} is relatively faster than ours', since they use simple linear SVMs, while we use deep CNNs and RNNs.

\section{Conclusion}
\label{SEC:Con}
In this paper we have presented a jointly trained network for simultaneous car license plate detection and recognition. With this network, car license plates can be detected and recognized all at once in a single forward pass, with both high accuracy and efficiency. By sharing convolutional features with both detection and recognition network, the model size decreases largely. The whole network can be trained approximately end-to-end, without intermediate processing like image cropping or character separation. Comprehensive evaluation and comparison on three datasets with different approaches validate the advantage of our method. In the future, we will extend our network to multi-oriented car license plates. In addition, with the time analysis, it is found that NMS takes about half of the whole processing time. Hence, we will optimize NMS to accelerate the processing speed.

\bibliographystyle{IEEEtran}
\bibliography{IEEEabrv,mybibfile}

\end{document}